\documentclass[11pt,a4paper]{article}
\usepackage[hyperref]{emnlp2018}
\usepackage{times}
\usepackage{amssymb}
\usepackage{amsmath}
\usepackage{latexsym}
\usepackage[font=footnotesize,figurewithin=none]{caption}
\usepackage{subcaption}
\usepackage{enumitem}
\usepackage{url}
\usepackage{graphicx}
\usepackage{color}
\usepackage{colortbl}
\usepackage{float}
\usepackage{todonotes}
\usepackage{multirow}%

\definecolor{mygrey}{gray}{0.85}

\def\argmax{\mathop{\rm argmax}}

\usepackage{verbatim}
\newcommand{\nr}[3]{$\langle#1 ;\, #2 ;\, #3\rangle$}

\aclfinalcopy 
\def\aclpaperid{1998} 

\newcommand\Mark[1]{\textsuperscript#1}

\title{Translating Navigation Instructions in Natural Language\\ to a High-Level Plan for Behavioral Robot Navigation}

\author{Xiaoxue Zang\Mark{1}\Thanks{ Both authors contributed equally to this work.}, Ashwini Pokle\Mark{1}\footnotemark[1], Marynel V\'azquez\Mark{2}, Kevin Chen\Mark{1}, \\ 
{\bf Juan Carlos Niebles\Mark{1}}, {\bf Alvaro Soto\Mark{3}}, {\bf Silvio Savarese\Mark{1}} \\
\Mark{1} Stanford University, \Mark{2} Yale University, \Mark{3} P. Universidad Cat\'olica de Chile \\
\Mark{1}\{xzang, 
ashwinipokle, kchen92, jniebles, ssilvio\}@stanford.edu, \\ 
\Mark{2}marynel.vazquez@yale.edu, \Mark{3}asoto@ing.puc.cl}
\date{}
\begin{document}

\setlength{\abovedisplayskip}{3pt}
\setlength{\belowdisplayskip}{3pt}

\maketitle
\begin{abstract}
We propose an end-to-end deep learning model for translating free-form natural language instructions to a high-level plan for behavioral robot navigation. The proposed model uses attention mechanisms to connect information from user instructions with a topological representation of the environment. To evaluate this model, we collected a new dataset for the translation problem containing 11,051 pairs of user instructions and navigation plans. Our results show that the proposed model outperforms baseline approaches on the new dataset. Overall, our work suggests that a topological map of the environment can serve as a relevant knowledge base for translating natural language instructions into a sequence of navigation behaviors.

\end{abstract}

\section{Introduction}

Enabling robots to follow navigation instructions in natural language can facilitate human-robot interaction across a variety of applications. For instance, within the service robotics domain, robots can follow navigation instructions to help with mobile manipulation \cite{tellex2011understanding} and
delivery tasks \cite{veloso2015cobots}. 

\begin{figure*}[tb]
\centering
\includegraphics[width=\linewidth]{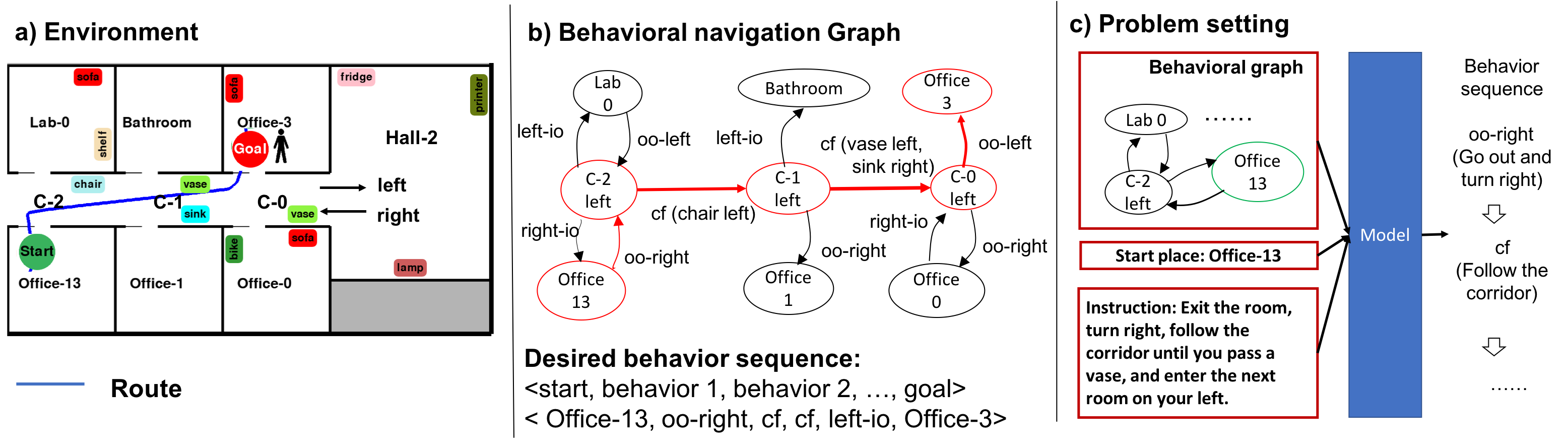}
\caption{
Map of an environment (a), its (partial) behavioral navigation graph (b), and the problem setting of interest (c). The  
red part of (b) corresponds to the representation of the route highlighted in blue in (a). The codes ``oo-left'',
``oo-right'', ``cf'', ``left-io'', and ``right-io'' correspond 
to the behaviors ``go out and turn left'', ``go out and turn right'', ``follow the corridor'', ``enter the room on left'', and ``enter office on right'', respectively. 
}
\label{fig:header}
\end{figure*}

Interpreting navigation instructions in natural language is difficult due to the high variability in the way people describe routes \cite{Chen:2011}. For example, there are a variety of ways to describe the route in Fig. \ref{fig:header}(a):
\begin{itemize}[align=left,labelsep=0em,noitemsep,topsep=0.2pt]
\item[--] \textit{``Exit the room, turn right, follow the corridor until you pass a vase on your left, and enter the next room on your left''}; or
\item[--] \textit{``Turn right after you exit the room, and enter the room on the left right before the end of the corridor''}; or
\item[--] \textit{``Advance forward to the right after going out of the door. Enter the room which is in the middle of two vases on your left.''}
\end{itemize}
Each fragment of a sentence within these instructions can be mapped to one or more than one navigation behaviors. For instance, assume
that a robot counts with a number of primitive, navigation behaviors, such as \textit{``enter the room on the left (or on right)''} , \textit{``follow the corridor''}, \textit{``cross the intersection''}, etc. Then, the fragment \textit{``advance forward''} in a navigation instruction could be interpreted as a \textit{``follow the corridor''} behavior, or as a sequence of \textit{``follow the corridor''} interspersed with \textit{``cross the intersection''} behaviors depending on the topology of the environment. Resolving such ambiguities often requires reasoning about ``common-sense'' concepts, as well as interpreting spatial information and landmarks, e.g., in sentences such as \textit{``the room on the left right before the end of the corridor''} and  \textit{``the room which is in the middle of two vases''}.


In this work, we pose the problem of interpreting navigation instructions as finding a mapping (or grounding) of the commands into an executable navigation plan. 
While the plan is typically modeled as a formal specification of low-level motions \cite{Chen:2011} or a  grammar \cite{artzi2013weakly, Matuszek:2010}, 
we focus specifically on translating instructions to a high-level navigation plan based on a topological representation of the environment. This representation is a \textit{behavioral navigation graph}, as recently proposed by \cite{Sepulveda:EtAl:2018}, designed to take advantage of the semantic structure typical of human environments. The nodes of the graph correspond to semantically meaningful locations for the navigation task, such as kitchens or entrances to rooms in corridors. The edges are parameterized, visuo-motor behaviors that allow a robot to navigate between neighboring nodes, as illustrated in Fig. \ref{fig:header}(b). Under this framework, complex navigation routes can be achieved by sequencing behaviors without an explicit metric representation of the world.


We formulate the problem of following instructions under the framework of \cite{Sepulveda:EtAl:2018} as finding a path in the behavioral navigation graph that follows the desired route, given a known starting location. The edges (behaviors) along this path serve to reach the -- sometimes implicit -- destination requested by the user. As in \cite{Zang18:inproceedings}, our focus is on the problem of interpreting navigation directions. We assume that a robot can realize valid navigation plans according to the graph.

We contribute a new end-to-end model for following directions in natural language under the behavioral navigation framework. Inspired by the information retrieval and question answering literature \cite{lewis1996natural,seo2016,xiong2016,Palangi:2016}, 
we propose to leverage the behavioral graph as a knowledge base to facilitate the interpretation of navigation commands. More specifically, the proposed model takes as input user directions in text form, the behavioral graph of the environment encoded as \texttt{$\langle$node;\,edge;\,node$\rangle$} triplets, and the initial location of the robot in the graph. The model then predicts a set of behaviors to reach the desired destination according to the instructions and the map (Fig.~\ref{fig:header}(c)). Our main insight is that using attention mechanisms to correlate navigation instructions with the topological map of the environment can facilitate predicting correct navigation plans.


This work also contributes a new dataset of $11,050$ pairs of free-form natural language instructions and high-level navigation plans. This dataset was collected through Mechanical Turk using 100 simulated environments with a corresponding topological map and, to the best of our knowledge, it is the first of its kind for behavioral navigation. The dataset opens up opportunities to explore data-driven methods for grounding navigation commands into high-level motion plans. 

We conduct extensive experiments to study the generalization capabilities of the proposed model for following natural language instructions. We investigate both generalization to new instructions in known and in new environments. We conclude this paper by discussing the benefits of the proposed approach as well as opportunities for future research based on our findings.

\section{Related work}

This section reviews relevant prior work on following navigation instructions. Readers interested in an in-depth 
review of methods to interpret spatial natural language for robotics are encouraged to refer to \cite{landsiedel2017review}.


Typical approaches to follow navigation commands deal with the complexity of natural language by manually parsing commands, constraining language descriptions, or using statistical machine translation methods. While manually parsing commands is often impractical, the first type of approaches are foundational: they showed that it is possible to leverage the compositionality of semantic units to interpret spatial language \cite{bugmann2004corpus, levit2007interpretation}. 

Constraining language descriptions can reduce the size of the input space to facilitate the interpretation of user commands. 
For example, \cite{FindMO:2016} explored using structured, symbolic language phrases for navigation. As in this earlier work, we are also interested in navigation with a topological map of the environment. However, we do not process symbolic phrases. Our aim is to translate free-form natural language instructions to a navigation plan using information from a high-level representation of the environment. This translation problem requires dealing with missing actions in navigation instructions and actions with preconditions, such as \textit{``at the end of the corridor, turn right''} \cite{MacMahon2006WalkTT}.

Statistical machine translation \cite{koehn2009statistical} is at the core of recent approaches to enable robots to follow navigation instructions. These methods aim to automatically discover translation rules from a corpus of data, and often leverage the fact that navigation directions are composed of sequential commands. For instance, \cite{wong2006learning, Matuszek:2010, Chen:2011} used statistical machine translation to map instructions to a formal language defined by a grammar. Likewise, \cite{kollar10:inproceedings, tellex2011understanding} mapped commands to spatial description clauses based on the hierarchical structure of language in the navigation problem. Our approach to  machine translation builds on insights from these prior efforts. In particular, we focus on end-to-end learning for statistical machine translation due to the recent success of Neural Networks in Natural Language Processing \cite{Goodfellow-et-al-2016}.


Our work is inspired by methods that reduce the 
task of interpreting user commands to a sequential prediction problem \cite{Shimizu2009LearningTF, mei:16, spoon:2018}. 
%
%
%
Similar to ~\citeauthor{mei:16} and ~\citeauthor{spoon:2018}, we use a \textit{sequence-to-sequence} model to enable a mobile agent to follow routes. But instead leveraging visual information to output low-level navigation commands, 
we focus on using a topological map of the environment to output a high-level navigation plan. This plan is a  sequence of behaviors that can be executed by a robot to reach a desired destination \cite{Sepulveda:EtAl:2018, Zang18:inproceedings}.

We explore machine translation from the perspective of automatic question answering. Following \cite{seo2016,xiong2016}, our approach uses attention mechanisms to learn alignments between different input modalities. In our case, the inputs to our model are navigation instructions, a topological environment map, and the start location of the robot (Fig.~\ref{fig:header}(c)). Our results show that the map can serve as an effective source of contextual information for the translation task. Additionally, it is possible to leverage this kind of information in an end-to-end 
fashion.

\section{Problem Formulation}
Our goal is to translate navigation instructions in text form into a sequence of behaviors that a robot can execute to reach  a desired destination from a known start location. We frame this problem under a behavioral approach to indoor autonomous navigation \cite{Sepulveda:EtAl:2018} and assume that prior knowledge about the environment is available for the translation task. This prior knowledge is a topological map, in the form of a behavioral navigation graph (Fig. \ref{fig:header}(b)). The nodes of the graph correspond to semantically-meaningful locations for the navigation task, and its directed edges are visuo-motor behaviors that a robot can use to move between nodes. This formulation takes advantage of the 
rich semantic structure behind man-made environments, resulting in a compact route representation for robot navigation.
 
Fig.~\ref{fig:header}(c) provides a schematic view of the problem setting. 
The inputs are:
(1) a navigation graph $m$, (2) the starting node $s$ of the robot in $m$, and (3) a set of free-form navigation instructions $I$ in natural language. The instructions describe a path in the graph to reach from $s$ to a -- potentially implicit -- destination node $g$. Using this information, the objective is to predict a suitable sequence of robot behaviors $b_1, \dots,b_T$ to navigate from $s$ to $g$ according to  $I$. From a supervised learning perspective, the goal is then to estimate:
\begin{equation}
\argmax_{b_1, \dots,b_T}  P(b_1, \dots,b_T | m, s, I)
\end{equation}

\noindent 
based on a dataset of input-target pairs $\{(\mathbf{x}_i,\, \mathbf{y}_i) \,|\, 0\leq i \leq N\}$, where $\mathbf{x}_i = (m, s, I)_i$ and $\mathbf{y}_i = (b_1, \dots,b_T)_i$, respectively.
The sequential execution of the behaviors $b_1, \dots,b_T$ should  replicate the route intended by the instructions $I$.

We assume no prior linguistic knowledge. Thus, translation approaches have to cope with the semantics and syntax of the language by discovering corresponding patterns in the data. 




\subsection{The Behavioral Graph: A Knowledge Base For Navigation}

We view the behavioral graph $m$ as a knowledge base that encodes a set of navigational rules as triplets \nr{p_i}{b_{l}[attr]}{p_j}, where 
$p_i$ and $p_j$ are adjacent nodes in the graph, and the edge $b_{l}$ is an executable behavior to navigate from $p_i$ to $p_j$. In general, each behaviors includes a list of relevant navigational attributes $attr$ that the robot might encounter when moving between nodes. 

\begin{table}[t]
\centering
\small{
\begin{tabular}{|c|c|}
\hline \bf Behavior & \bf Description \\ \hline
oo$<$d$>$ & Go out of the current place and turn $<$d$>$\\ \arrayrulecolor{mygrey}\hline\arrayrulecolor{black}
io$<$d$>$  & Turn $<$d$>$ and enter the place straight ahead \\ \arrayrulecolor{mygrey}\hline\arrayrulecolor{black}
oio & Exit current place and enter straight ahead \\ \arrayrulecolor{mygrey}\hline\arrayrulecolor{black}
$<$d$>$t & Turn $<$d$>$ at the intersection \\ \arrayrulecolor{mygrey}\hline\arrayrulecolor{black}
cf & Follow (or go straight down) the corridor \\ \arrayrulecolor{mygrey}\hline\arrayrulecolor{black}
sp & Go straight at a T intersection \\ \arrayrulecolor{mygrey}\hline\arrayrulecolor{black}
ch$<$d$>$ & Cross the hall and turn $<$d$>$ \\
\hline
\end{tabular}
}
\caption{\label{tab:act} Behaviors (edges) of the  navigation graphs considered in this work. The direction $<$d$>$ can be left or right. }
\vspace{-1em}
\end{table}

We consider 7 types of semantic locations, 11 types of behaviors, and 20 different types of landmarks. A location in the navigation graph can be a room, a lab, an office, a kitchen, a hall, a corridor, or a bathroom. These places are labeled with unique tags, such as "room-1" or "lab-2", except for bathrooms and kitchens which people do not typically refer to by unique names when describing navigation routes. 

Table \ref{tab:act} lists the navigation behaviors that we consider in this work. These behaviors can be described in reference to visual landmarks or objects, such as paintings, book shelfs, tables, etc. As in Fig.~\ref{fig:header}, maps might contain multiple landmarks of the same type. Please see the supplementary material (Appendix A) for more details.

\section{Approach}\label{sec:modal}
\begin{figure*}[tb]
\centering
\includegraphics[width=0.7\linewidth]{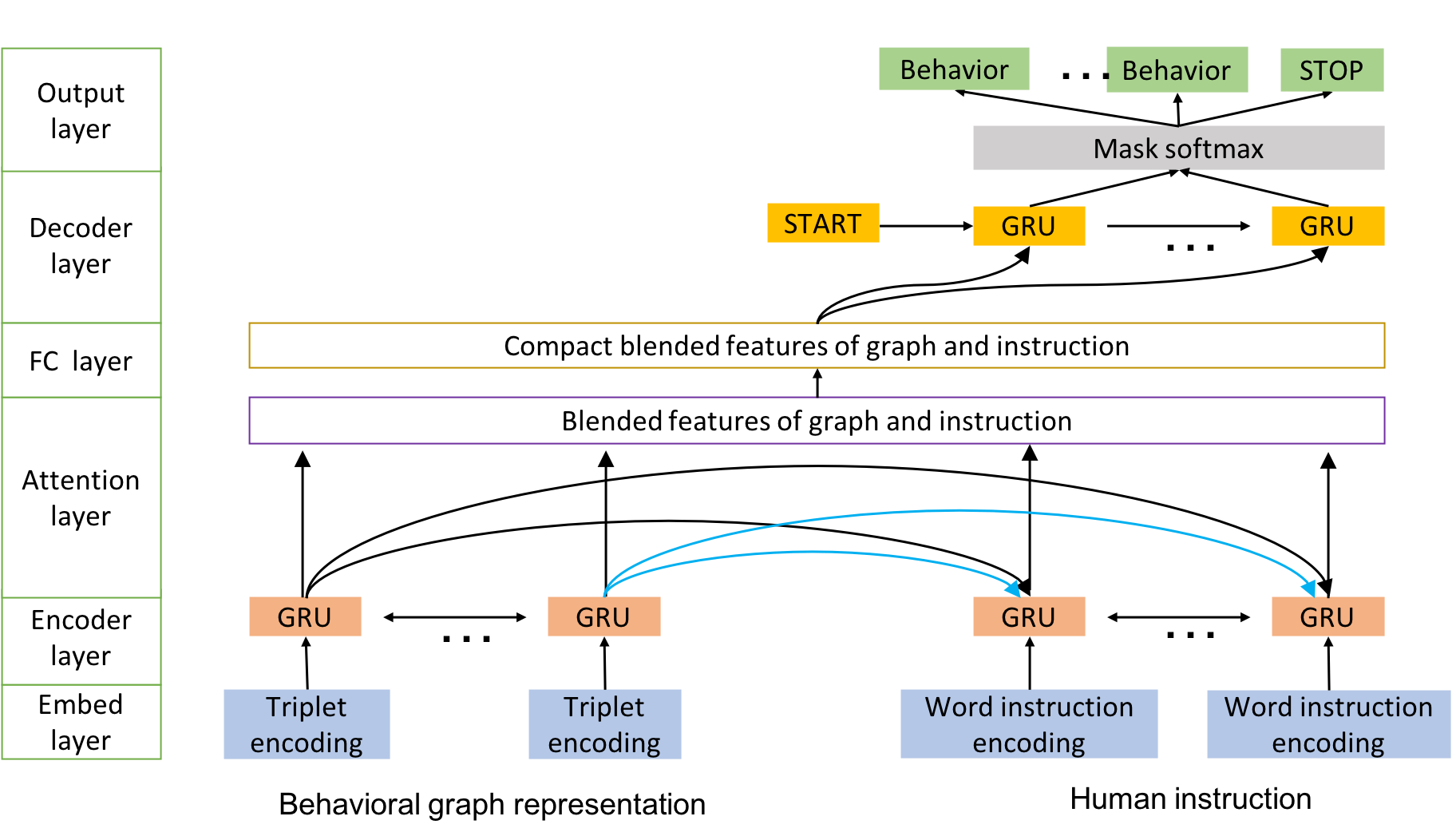}
\caption{Model overview. The model contains six layers, takes the input of behavioral graph representation, free-form instruction, and the start location (yellow block marked as START in the decoder layer) and outputs a sequence of behaviors.}
\label{fig:model}
\end{figure*}
We leverage recent advances in deep learning to translate natural language instructions to a sequence of navigation behaviors in an end-to-end fashion. Our proposed model builds on the sequence-to-sequence translation model of \cite{bah2015}, which computes a 
soft-alignment between a source sequence (natural language instructions in our case) and the corresponding target sequence (navigation behaviors). 

As one of our main contributions, we augment the neural machine translation approach of \citeauthor{bah2015} to take as input not only  natural language instructions, but also the corresponding 
 behavioral navigation graph $m$ of the environment where navigation should take place. Specifically, at each step, the graph $m$ operates as a knowledge base that the model can access to obtain information about path 
connectivity, facilitating the grounding of navigation commands.

Figure \ref{fig:model} shows the structure of the proposed model for interpreting navigation instructions. The model consists of six layers:

{\bf Embed layer}: The model first encodes each word and symbol in the input sequences $I$ and $m$ into fixed-length representations. The instructions $I$ are embedded into a 100-dimensional pre-trained GloVe vector \citep{glove}. Each of the triplet components, $p_i$, $b_l[attr]$, and $p_j$ of the graph $m$, are one-hot encoded into vectors of dimensionality $2N + E$, where $N$ and $E$ are the number of nodes and edges in $m$, respectively. 

{\bf Encoder layer}: The model then uses two bidirectional Gated Recurrent Units (GRUs)~\cite{gru} to independently process the information from 
$I$ and 
$m$, and incorporate contextual cues from the surrounding embeddings in each sequence. The outputs of the encoder layer are the matrix $\bar{I} \in 
\mathbb{R}^{T\times 2H}$ for the navigational commands and the  matrix $\bar{G} \in \mathbb{R}^{L\times 2H}$ for the behavioral graph, where $H$ is the hidden size of each GRU, $T$ is the number of words in the instruction $I$, and $L$ is the number of triplets in the graph $m$.


{\bf Attention layer}:
Matrices $\bar{I}$ and $\bar{G}$ generated by the encoder layer are combined using an attention mechanism. We use one-way attention because the graph contains information about the whole environment, while the instruction has (potentially incomplete) local information about the route of interest. The use of attention 
provides our model with a two-step strategy to interpret commands. This resembles the way people find paths on a map: first, relevant parts on the map are selected according to their affinity to each of the words in the input instruction (attention 
layer); second, the selected parts are connected to assemble a valid path (decoder 
layer). More formally, let $\bar{G}_i$ ($i \in [1,L]$) be the $i$-th row of $\bar{G}$, and $\bar{I}_j$ ($j \in [1,T]$) the $j$-th row of $\bar{I}$. 
We use each encoded triplet $\bar{G}_i$ in $\bar{G}$ to calculate its associated attention
distribution $a_i \in \mathbb{R}^T$ over all the atomic instructions $\bar{I}_j$: 
\begin{align} 
e_i &= [\bar{G}_i W \bar{I}_1^\intercal, \dots , \bar{G}_i W\bar{I}_T^\intercal ]\\
a_i &= softmax(e_i) 
\end{align} 

\noindent 
where the matrix $W \in \mathbb{R}^{2H\times 2H}$ 
serves 
to combine the 
different sources of information $\bar{G}$ and $\bar{I}$. 
%
Each component 
$a_{ij}$ of the attention distributions $a_i$ quantifies the affinity between the $i$-th triplet in $\bar{G}$ and the $j$-th word in the corresponding input
$I$. 

The model then uses each attention 
distribution $a_i$ to obtain a weighted sum of the encodings of the words in $\bar{I}$, according to their relevance to the corresponding triplet $\bar{G_i}$. This results in L 
attention vectors $R_i \in \mathbb{R}^{2H}$, $R_i = \sum_{j=1}^{T} a_{ij}I_j$.

The final step in the attention layer concatenates each $R_i$ with $\bar{G}_i$ to generate the outputs $F_i= [R_i; 
\bar{G}_i]$, $i \in [1,L]$.
%
Following \cite{seo2016}, we  include the 
encoded triplet $\bar{G}_i$ in the output tensor $F_i$ of this layer to prevent early summaries of relevant map information. 

{\bf FC layer}: The model reduces the dimensionality of each individual 
vector $F_i$ from $4H$ to $H$ with a fully-connected (FC) layer. The resulting L vectors are  output to the next layer as columns of a context matrix $C \in \mathbb{R}^{H \times L}$.

{\bf Decoder layer}: After the FC layer, the model predicts likelihoods over the sequence of behaviors that correspond to the input instructions with a GRU network. Without loss of generality, consider the $t$-th recurrent cell in the GRU network. This cell takes two inputs: a hidden state vector $h_{t-1}$ from the prior cell, and a one-hot embedding of the previous behavior $b_{t-1}$ that was predicted by the model. Based on these inputs, the GRU cell outputs a new hidden state $h_{t}$ to compute likelihoods for the next behavior. These likelihoods are estimated by combining the output state $h_{t}$ with relevant information from the context $C$:
\begin{align}\label{eq:bah1}
\hat{d}_{ts}&= v_a^\intercal \tanh(W_1 h_t + W_2 C_s) \\
d_{t} &= softmax(\hat{d}_{t1}, \dots, \hat{d}_{tL}) \label{eq:decoderAttention}
\end{align}
where $W_1$, $W_2$, and $v_a$ are trainable parameters. The attention vector $d_{t} 
\in \mathbb{R}^{L}$ in Eq. (\ref{eq:decoderAttention})  quantifies the affinity of $h_t$ with respect to each of the columns $C_s$ of $C$, where $s \in [1,L]$. 
The attention vector also helps to estimate a dynamic contextual vector $S_t = \sum_{s=1}^{L} d_{ts}C_s$ that the $t$-th GRU cell uses to compute logits 
for the next behavior:
\begin{equation}
o_t = W_3[S_t;h_t]
\label{eq:logits}
\end{equation}
with $W_3$ trainable parameters. Note that $o_t$ includes a value for each of the pre-defined behaviors in the graph $m$, as well as for a special ``\textit{stop}'' symbol to identify the end of the output sequence. 

{\bf Output layer}: The final layer of the model searches for a valid sequence of robot behaviors based on the robot's initial node, the connectivity of the graph $m$, and the output logits from the previous decoder layer. 
%
Again, without loss of generality, consider the $t$-th behavior $b_t$ that is finally predicted by the model. The search for this behavior is implemented as:
\begin{equation}
b_t=argmax(softmax(o_t + mask(m, n_{t}))) 
\label{eq:mask}
\end{equation}
with $mask(m, n_{t})$ a masking function that takes as input the graph $m$ and the node $n_{t}$ that the robot reaches after following the sequence of behaviors $b_1, \ldots, b_{t-1}$ previously predicted by the model. The $mask$ function returns a vector of the same dimensionality as the logits $o_t$, but with zeros for the valid behaviors after the last location $n_{t}$ and for the special stop symbol, and $-\inf$ for any invalid predictions according to the connectivity of the behavioral navigation graph.

\section{Dataset}\label{sec:dataset}
We created a new dataset for the problem of following navigation instructions under the behavioral navigation framework of \cite{Sepulveda:EtAl:2018}.\footnote{The dataset is publicly available through the website: \url{follow-nav-directions.stanford.edu.}} This dataset was created using Amazon Mechanical Turk and 100 maps of simulated indoor environments, each with 6 to 65 rooms. To the best of our knowledge, this is the first benchmark for comparing translation models in the context of behavioral robot navigation.




As shown in Table \ref{tab:data}, the dataset consists of 8066 pairs of free-form natural language instructions and navigation plans for training. This training data was collected from 88 unique simulated environments, totaling 6064 distinct navigation plans (2002 plans have two different navigation instructions each; the rest has one). The dataset contains two test set variants:
\begin{description}[align=left,leftmargin=0em,labelsep=0.2em,font=\textbf]
\item[1) Test-Repeated:] Contains 1012 pairs of instructions and navigation plans. These routes are not part of the training set; however, they are collected using environments that are part of the training set. 
\item[2) Test-New:] Contains 962 pairs of instructions and navigation plans. This test set is more challenging than the Test-Repeated dataset because it contains new routes on 12 new indoor environments not included in the training set.
\end{description}
While the dataset was collected with simulated environments, no structure was imposed on the navigation instructions while crowd-sourcing data. Thus, many instructions in our dataset are ambiguous. Moreover, the order of the behaviors in the instructions is not always the same. For instance, a person said \textit{``turn right and advance''} to describe part of a route, while another person said \textit{``go straight after turning right''} in a similar situation. The high variability present in the natural language descriptions of our dataset makes the problem of decoding instructions into behaviors not trivial.
See Appendix A of the supplementary material for additional details on our data collection effort.  

\begin{table}[t]
\centering
\small{
\begin{tabular}{|c|c|c|c|}
\hline 
\bf Dataset & \bf \# Single & \bf \# Double & \bf Total \\
\hline
Training & 4062 & 2002 & 8066 \\
\arrayrulecolor{mygrey}\hline\arrayrulecolor{black}
Test-Repeated 
&  944 & 34 & 1012 \\
\arrayrulecolor{mygrey}\hline\arrayrulecolor{black}
Test-New 
&  962 & 0 & 962 \\
\hline
\end{tabular}
}
\caption{ Dataset statistics. ``\# Single'' indicates  the number of navigation plans with a single natural language instruction. ``\# Double'' is the number of plans with two different instructions. The total number of plans is (\# Single) $\times$ 2(\# Double).}
\label{tab:data}
\end{table}



\section{Experiments}
This section describes our evaluation of the proposed approach for interpreting navigation commands in natural language. We provide both quantitative and qualitative results.

\subsection{Evaluation Metrics}\label{sec:metrics}
While computing evaluation metrics, we only consider the behaviors present in the route because they are sufficient to recover the high-level navigation plan from the graph. Our metrics treat each behavior as a single token. For example, the sample plan ``R-1 oor C-1 cf C-1 lt C-0 cf C-0 iol O-3" is considered to have 5 tokens, each corresponding to one of its behaviors (``oor", ``cf", ``lt", ``cf", ``iol"). In this plan, ``R-1",``C-1", ``C-0", and ``O-3" are symbols for  locations (nodes) in the graph.

We compare the performance of translation approaches based on four metrics:
\begin{description}[align=left,leftmargin=0em,labelsep=0.4em,font=\textbf]
\item[- Exact Match (EM).] As in \cite{Shimizu2009LearningTF}, EM is 1 if a predicted plan matches exactly the ground truth; otherwise it is 0.
\item[- F1 score (F1).] The harmonic average of the precision and recall over all the test set \cite{chinchor1993muc}.
\item[- Edit Distance (ED).] The minimum number of insertions, deletions or swap operations required to transform a predicted sequence of behaviors into the ground truth sequence \cite{navarro2001guided}.
\item[- Goal Match (GM).] GM is 1 if a predicted plan reaches the ground truth destination (even if the full sequence of behaviors does not match exactly the ground truth). Otherwise, GM is 0. 

\end{description}




\subsection{Models Used in the Evaluation}\label{sec:models-eval}
We compare the proposed approach for translating natural language instructions into a navigation plan against alternative deep-learning models:
\begin{description}[align=left,leftmargin=0em,labelsep=0.4em,font=\textbf]
\item[Baseline model.]
The baseline approach is based on \cite{Shimizu2009LearningTF}. It divides the task of interpreting commands for behavioral navigation into two steps: path generation, and path verification. For path generation, this baseline uses a standard sequence-to-sequence model augmented with an attention mechanism, similar to \cite{bah2015,Zang18:inproceedings}. For path verification, the baseline uses depth-first search to find a route in the graph that matches the sequence of predicted behaviors. If no route matches perfectly, the baseline changes up to three behaviors in the predicted sequence to try to turn it into a valid path.



\item[Ablation model.]
To test the impact of using the behavioral graphs as an extra input to our translation model, we implemented a version of our approach that only takes natural language instructions as input. In this ablation model, the output of the bidirectional GRU that encodes the input instruction $I$ is directly fed to the decoder layer. This model does not have the attention and FC layers described in Sec. \ref{sec:modal}, nor uses the masking function in the output layer.

\item[Ablation with mask model.] This model is the same as the previous Ablation model, but with the masking function in the output layer.
\end{description}

%
%

\subsection{Implementation Details}\label{sec:exp}
We pre-processed the inputs to the various models that are considered in our experiment. In particular, we lowercased, tokenized, spell-checked and lemmatized the input instructions in text-form using WordNet~\cite{wordnet}. We also truncated the graphs to a maximum of 300 triplets, and the navigational instructions to a maximum of 150 words. Only 6.4\% (5.4\%) of the unique graphs in the training (validation) set had more than 300 triplets, and less than 0.15\% of the natural language instructions in these sets had more than 150 tokens.

The dimensionality of the hidden state of the GRU networks was set to 128 in all the experiments. In general, we used 12.5\% of the training set as validation for choosing models' hyper-parameters. In particular, we used dropout after the encoder and the fully-connected layers of the proposed model to reduce overfitting. Best performance was achieved with a dropout rate of 0.5 and batch size equal to 256. We also used scheduled sampling~\cite{bengio2015scheduled} at training time for all models except the baseline.

We input the triplets from the graph to our proposed model in alphabetical order, and consider a modification where the triplets that surround the start location of the robot are provided first in the input graph sequence. We hypothesized that such rearrangement would help identify the starting location (node) of the robot in the graph. In turn, this could facilitate the prediction of correct output sequences. In the remaining of the paper, we refer to models that were provided a rearranged graph, beginning with the starting location of the robot, as models with ``Ordered Triplets''. 

\subsection{Quantitative Evaluation}\label{sec:quant}

\begin{table*}[t!]
\centering
\small{
\begin{tabular}{ccccc|cccc}
\multirow{ 2}{*}{\bf Model} & \multicolumn{4}{c|}{\bf Test-Repeated Set} & \multicolumn{4}{c}{\bf Test-New Set} \\
 & \bf EM $\uparrow $ & \bf F1 $\uparrow $ & \bf ED $\downarrow$ & \bf GM $\uparrow $& \bf EM $\uparrow $ & \bf F1$\uparrow $ & \bf ED $\downarrow$& \bf GM $\uparrow $ \\
\hline
Baseline & 25.30 & 79.83  & 2.53 & 26.28 & 25.44 & 81.38 & 2.39	& 25.44\\
\arrayrulecolor{mygrey}\hline\arrayrulecolor{black}
Ablation & 36.36 & 90.28 & 1.36 & 36.36 & 24.82 & 88.65 & 1.71	& 24.92\\
\arrayrulecolor{mygrey}\hline\arrayrulecolor{black}
Ablation with Mask & 45.95 & 90.08 & 1.20 & 46.05 & 36.45 &	88.31 &	1.45 & 36.56\\
\arrayrulecolor{mygrey}\hline\arrayrulecolor{black}
Ours without Mask & 52.47 & 91.74 & 0.95 &	53.95 & 21.94 & 87.50 & 1.78 & 22.65 \\
\arrayrulecolor{mygrey}\hline\arrayrulecolor{black}
Ours with Mask & 57.31 &	91.91 &	0.91 &	57.31 & 38.52 & 88.98 & 1.32	& 38.52\\
\arrayrulecolor{mygrey}\hline\arrayrulecolor{black}
Ours without Mask and with Ordered Triplets & 57.21 &	93.37 &	0.79 &	57.71 & 33.36 & \textbf{91.02} & 1.37 & 33.78\\
\arrayrulecolor{mygrey}\hline\arrayrulecolor{black}
Ours with Mask and Ordered Triplets & \textbf{61.17} & \textbf{93.54} & \textbf{0.75} & \textbf{61.36} & \textbf{41.71} & 90.22 & \textbf{1.22} & \textbf{41.81} \\
\end{tabular}
}
\caption{Performance of different models on the test datasets. EM and GM report percentages, and ED corresponds to average edit distance. The symbol $\uparrow $ indicates that higher results are better in the corresponding column; $\downarrow$ indicates that lower is better.}
\label{tab:results}
\vspace{-1em}
\end{table*}

Table \ref{tab:results} shows the performance of the models considered in our evaluation on both test sets. The next two sections discuss the results in detail.

\subsubsection{Performance in the Test-Repeated Set}
First, we can observe that the final model ``Ours with Mask and Ordered Triplets'' outperforms the Baseline and Ablation models on all metrics in previously seen environments. The difference in performance is particularly evident for the Exact Match and Goal Match metrics, with our model increasing accuracy by 35\% and 25\% in comparison to the Baseline and Ablation models, respectively.
These results suggest that providing the behavioral navigation graph to the model and allowing it to process this information as a knowledge base in an end-to-end fashion is beneficial.


We can also observe from Table \ref{tab:results} that the masking function of Eq. (\ref{eq:mask}) tends to increase performance in the Test-Repeated Set by constraining the output sequence to a valid set of navigation behaviors. For the Ablation model, using the masking function leads to about $10\%$ increase in EM and GM accuracy. For the proposed model (with or without reordering the graph triplets), the increase in accuracy is around  $4\%$. Note that the impact of the masking function is less evident in terms of the F1 score because this metric considers if a predicted behavior exists in the ground truth navigation plan, irrespective of its specific position in the output sequence.

The results in the last four rows of Table \ref{tab:results} suggest that ordering the graph triplets can facilitate predicting correct navigation plans in previously seen environments. Providing the triplets that surround the starting location of the robot first to the model leads to a  boost of $4\%$ in EM and GM performance. The rearrangement of the graph triplets also helps to reduce ED and increase F1.
 
Lastly, it is worth noting that our proposed model (last row of Table \ref{tab:results}) outperforms all other models in previously seen environments. In particular, we obtain over $4\%$ increase in EM and GM between our model and the next best two models.

\subsubsection{Performance in the Test-New Set}

The previous section evaluated model performance on new instructions (and corresponding navigation plans) for environments that were previously seen at training time. Here, we examine whether the trained models succeed on environments that are completely new.

The evaluation on the Test-New Set helps understand the generalization capabilities of the models under consideration. This experiment is more challenging than the one in the previous section, as can be seen in performance drops in Table~\ref{tab:results} for the new environments. Nonetheless, the insights from the previous section still hold: masking in the output layer and reordering the graph triplets tend to increase performance. 


Even though the results in Table~\ref{tab:results} suggest that there is room for future work on decoding natural language instructions, our model still outperforms the baselines by a clear margin in new environments. For instance, the difference between our model and the second best model in the Test-New set is about $3\%$ EM and GM. Note that the average number of actions in the ground truth output sequences is 7.07 for the Test-New set. Our model's predictions are just $1.22$ edits off on average from the 
correct navigation plans. 


\subsection{Qualitative Evaluation}
This section discusses qualitative results to better understand how the proposed model uses the navigation graph.

\subsubsection{Attention Visualization}\label{sec:dis}

\begin{figure}[tb]
\centering
\includegraphics[width=\linewidth]{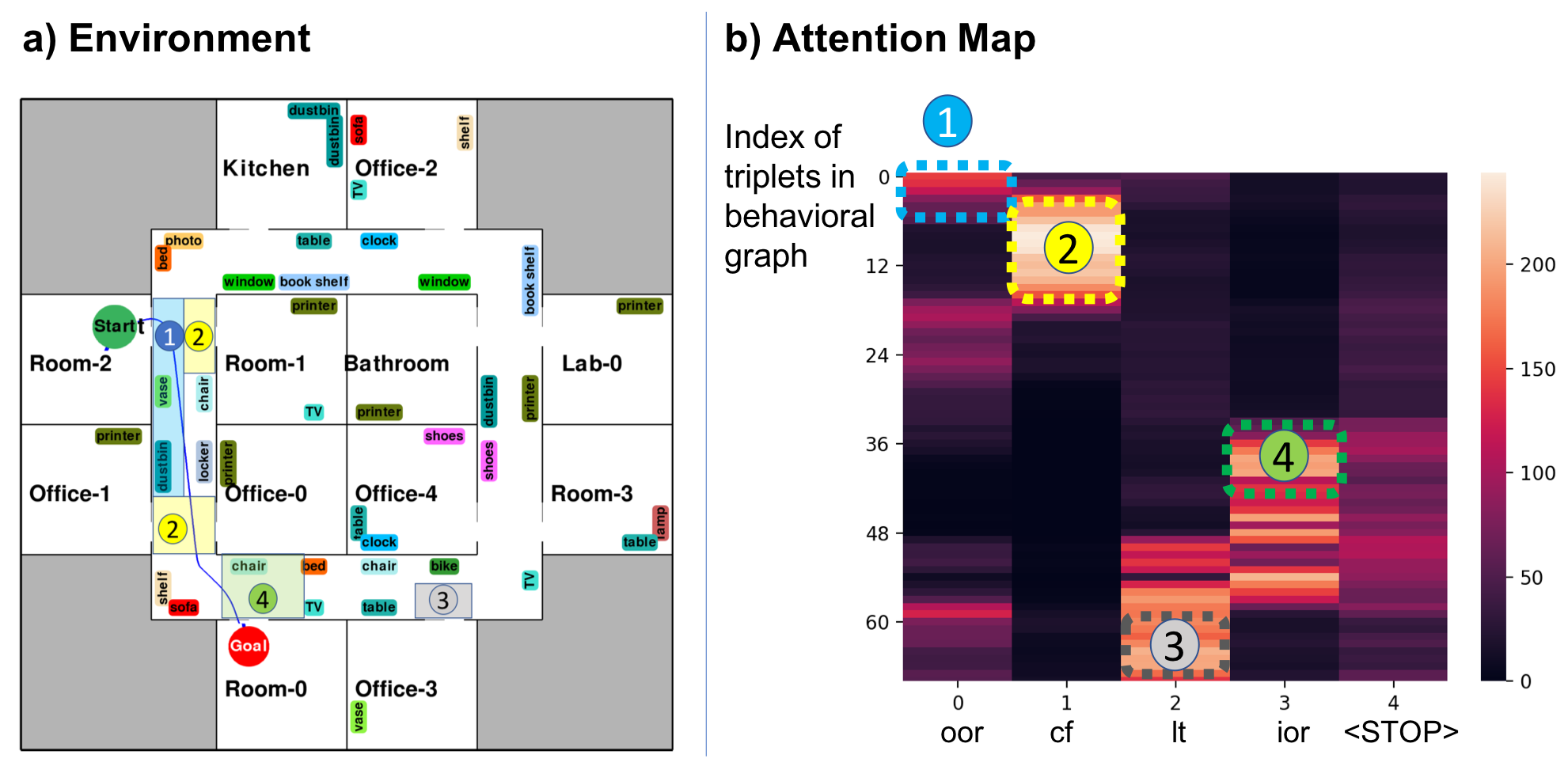}
\centering
\caption{Visualization of the attention weights of the decoder layer. The color-coded and numbered regions on the map (left) correspond to the triplets that are highlighted with the corresponding color in the attention map (right). }
\label{fig:attention}
\end{figure}

We analyze the evolution of the attention weights $d_t$ in Eq. (\ref{eq:decoderAttention}) to assess if the decoder layer of the proposed model is attending to the correct parts of the behavioral graph when making predictions. Fig~\ref{fig:attention}(b) shows an example of the resulting attention map for the case of a correct prediction. In the Figure, the attention map is depicted as a scaled and normalized 2D array of color codes. Each column in the array shows the attention distribution $d_t$ used to generate the predicted output at step $t$. Consequently, each row in the array represents a triplet in the corresponding behavioral graph. This graph consists of 72 triplets for Fig~\ref{fig:attention}(b).

We observe a locality effect associated to the attention coefficients corresponding to high values (bright areas) in each column of Fig~\ref{fig:attention}(b). This suggests that the decoder is paying attention to graph triplets associated to particular neighborhoods of the environment in each prediction step. 
%
We include additional attention visualizations in the supplementary Appendix, 
including cases where the dynamics of the attention distribution are harder to interpret.

\subsubsection{Experiments with Sub-Optimal Paths}

\begin{figure}[tb]
\centering
\includegraphics[width=0.8\linewidth]{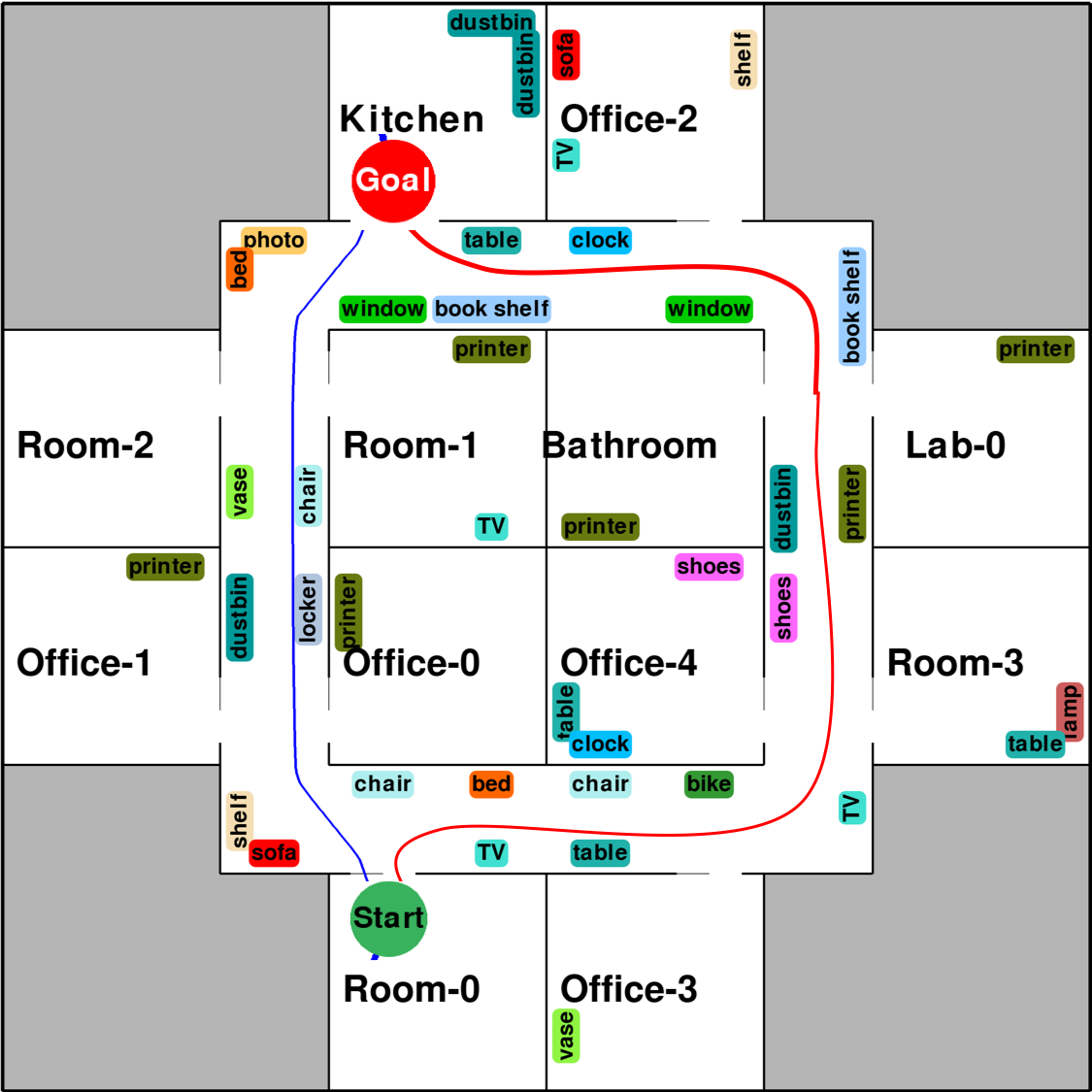}
\centering
\caption{An example of two different navigation paths between the same pair of start and goal locations.}
\label{fig:customize}
\end{figure}

All the routes in our dataset are the shortest paths from a start location to a given destination. Thus, 
we collected a few additional natural language instructions to check if our model was able to follow navigation instructions describing sub-optimal paths. 
%
One such example is shown in Fig.~\ref{fig:customize}, where the blue route (shortest path) and the red route (alternative path) are described by:
\begin{description}[leftmargin=*, labelsep=0.2em, itemsep=0em]
\item[-- Blue route:] ``\textit{Go out the office and make a left. Turn right at the corner and go down the hall. Make a right at the next corner and enter the kitchen in front of table.}''
\item[-- Red route:] ``\textit{Exit the room 0 and turn right, go to the end of the corridor and turn left, go straight to the end of the corridor and turn left again. After passing bookshelf on your left and table on your right, Enter the kitchen on your right.}''
\end{description}

For both routes, the proposed model was able to predict the correct sequence of navigation behaviors. This result suggests that the model is indeed using the input instructions and is not just approximating shortest paths in the behavioral graph. Other examples on the prediction of sub-obtimal paths are described in the Appendix.


\section{Conclusion}
This work introduced behavioral navigation through free-form natural language instructions as a challenging and a novel task that falls at the intersection of natural language processing and robotics. This problem has a range of interesting cross-domain applications, including information retrieval. 

We proposed an end-to-end system to translate user instructions to a high-level navigation plan. 
Our model utilized an attention mechanism to merge relevant information from the navigation instructions with a behavioral graph of the environment. The model then used a decoder to predict a sequence of navigation behaviors that matched the input commands.

As part of this effort, we contributed a new 
dataset of 11,051 pairs of user instructions and navigation plans from 100 different environments. Our model 
achieved the best performance in this dataset in comparison to a two-step baseline approach for interpreting navigation instructions, and a sequence-to-sequence model that does not consider the behavioral graph. Our quantitative and qualitative results suggest 
that attention mechanisms can help leverage the behavioral graph as a relevant knowledge base to facilitate the translation of free-form navigation instructions.
Overall, our approach demonstrated practical form of learning for a complex and useful task.

In future work, we are interested in investigating mechanisms to improve generalization to new environments. For example, pointer and graph networks~\cite{NIPS2015_5866, DefferrardBV16} are a promising direction to help supervise translation models and predict motion behaviors. 

\section*{Acknowledgments}
The Toyota Research Institute (TRI) provided funds to assist with this research, but this paper solely reflects the opinions and conclusions of its authors and not TRI or any other Toyota entity. This work is also partially funded by Fondecyt grant 1181739, Conicyt, Chile. The authors would also like to thank Gabriel Sep\'ulveda for his assistance with parts of this project.

\bibliography{emnlp2018}
\bibliographystyle{acl_natbib_nourl}
\end{document}